\documentclass[sigconf,balance=false]{acmart}

\setcopyright{none}
\copyrightyear{2027}
\acmYear{2027}
\acmConference[WWW '27]{The ACM Web Conference 2027}{May 10--14, 2027}{Dublin, Ireland}
\acmBooktitle{Proceedings of the ACM Web Conference 2027 (WWW '27)}
\acmDOI{}
\acmISBN{}
\usepackage{booktabs}
\usepackage{amsmath}
\usepackage{multirow}
\usepackage{colortbl}
\usepackage{pifont}
\usepackage{flafter}
\usepackage{placeins}
\usepackage{array}
\graphicspath{{fig/}}
\IfFileExists{algorithm.sty}{\usepackage{algorithm}}{}
\IfFileExists{algorithmic.sty}{\usepackage{algorithmic}}{}
\usepackage[T1]{fontenc}
\usepackage[utf8]{inputenc}
\title[Do Transaction Prices Reflect Visual Preference?]{Do Transaction Prices Reflect Visual Preference? A Diagnostic Study with Multimodal Language Models}

\author{Fengxian Ji}
\email{fengxian.ji@mbzuai.ac.ae}
\affiliation{%
  \institution{MBZUAI}
  \country{United Arab Emirates}}

\author{Jingpu Yang}
\email{jingpuyang290@gmail.com}
\affiliation{%
  \institution{Zhongguancun Academy}
  \city{Beijing}
  \country{China}}

\author{Fan Zhang}
\email{zhang.fan@mbzuai.ac.ae}
\affiliation{%
  \institution{MBZUAI}
  \country{United Arab Emirates}}

\author{Junhong Liang}
\email{junhong.liang@mbzuai.ac.ae}
\affiliation{%
  \institution{MBZUAI}
  \country{United Arab Emirates}}

\author{Qian Jiang}
\affiliation{%
  \institution{Northeastern University}
  \country{China}}

\author{Yuke Li}
\affiliation{%
  \institution{Northeastern University}
  \country{China}}

\author{Jinri Guo}
\affiliation{%
  \institution{Northeastern University}
  \country{China}}

\author{Rongxin Zhu}
\affiliation{%
  \institution{The University of Melbourne}
  \country{Australia}}

\author{Zhuohan Xie}
\authornote{Corresponding author.}
\email{zhuohan.xie@mbzuai.ac.ae}
\affiliation{%
  \institution{MBZUAI}
  \country{United Arab Emirates}}

\renewcommand{\shortauthors}{Ji et al.}

\begin{document}
\begin{abstract}
Existing visual preference evaluation relies on costly human annotations or preference reward models, whereas online marketplace transaction prices are readily available and reflect real purchasing behavior, offering potential low-cost supervision.
However, transaction prices conflate visual preferences with non-visual factors such as provenance, demand, and competition, making aggregate accuracy insufficient to distinguish model errors, unreliable price signals, and protocol confounds.
To address this identifiability challenge, we introduce \textbf{\textit{MarketImage}}, a diagnostic framework that identifies when transaction prices reflect visual preferences, where they fail, and whether multimodal language models can recover human-supported visual preference signals.
MarketImage comprises three components: 
(i) \textbf{\textit{MarketImageArt}}, 266 six-image comparison sets with transaction prices and price-blind human rankings; (ii) \textbf{\textit{MarketImageLens}}, a tie-preserving protocol that partitions 3,990 within-set image pairs into price–human agreement, conflict, and price-tie regions while separately auditing presentation-order effects; and (iii) \textbf{\textit{MarketValuator}}, six elicitation methods spanning direct comparison, independent scoring, budget constraints, and multi-agent aggregation to evaluate signal recovery by static visual scorers and three multimodal language models.
Experiments show that only 24.2\% of image pairs have a non-tied price direction, of which 34.7\% agree with the human majority. Within this region, HPSv2 leads at 66.4\% Recoverable Accuracy; Qwen's budgeted and aggregated variants trail baselines, while GPT interface gains remain statistically unresolved.
Thus, transaction prices encode recoverable visual preferences only under limited conditions and should be treated as uncertain proxies rather than ground truth.
Code: \href{https://github.com/FengxianJi/MarketImage}{GitHub}.
\end{abstract}

\ccsdesc[500]{Computing methodologies~Artificial intelligence~Computer vision}
\ccsdesc[300]{Human-centered computing~Human computer interaction ~Empirical studies in HCI}
\keywords{visual preference, transaction prices, multimodal large language models, human evaluation, benchmark}
\maketitle

\begin{figure}[t]
\centering
\includegraphics[width=\columnwidth]{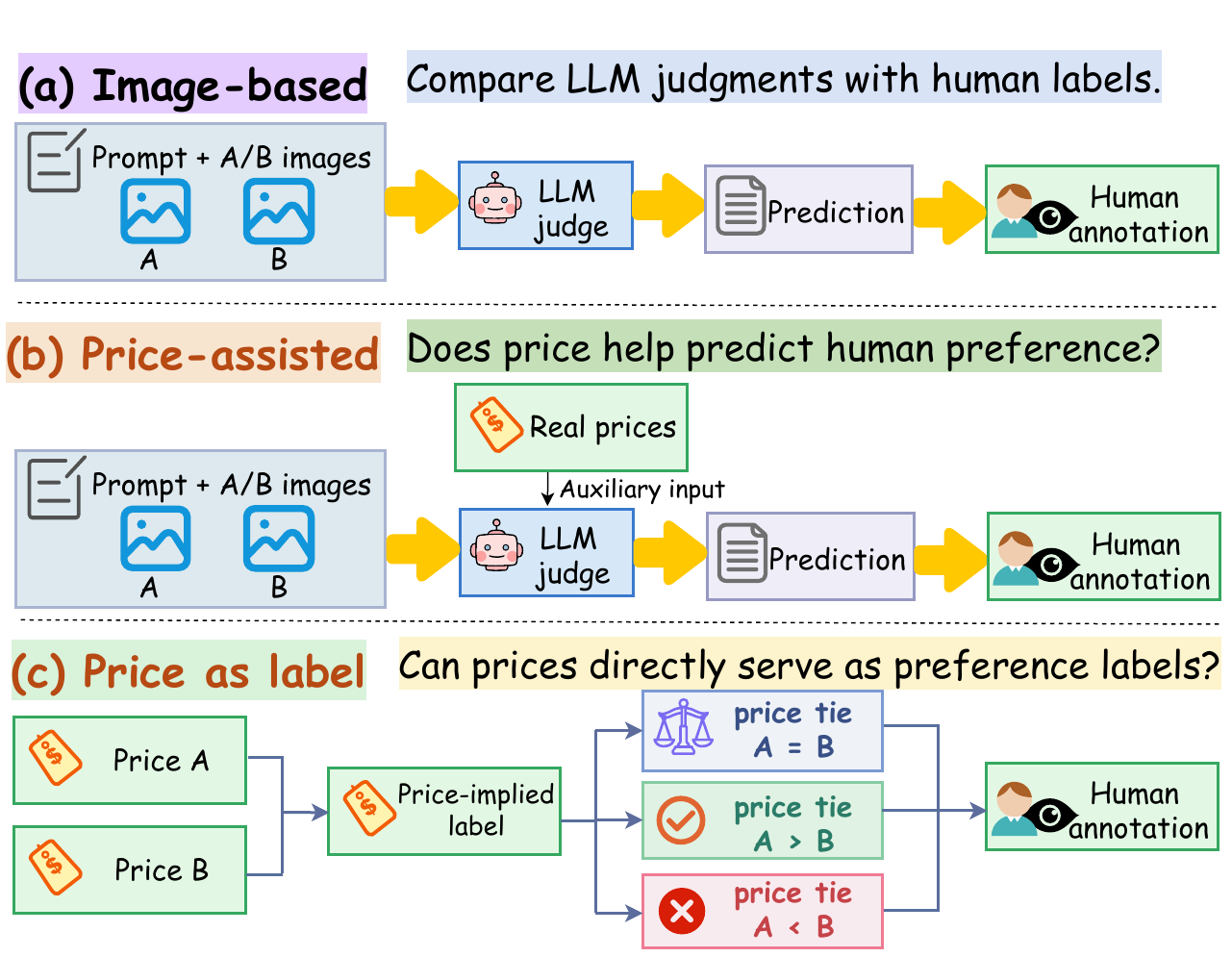}
\caption{
Conceptual progression from (a) LLM judgments evaluated against price-blind
human rankings, to (b) prior price-assisted prediction with human preferences
as targets, and (c) our test of transaction prices as candidate preference
labels. Panel (c) measures price coverage and agreement with human rankings
while preserving ties and conflicts. Prices are treated as a
protocol-dependent proxy, not assumed to be ground truth.
}
\label{fig:compared-with-cheaptalk}
\end{figure}

With the growing adoption of generative vision models in practical applications, evaluating whether generated images meet human needs and preferences has become an important research problem~\cite{li2024genai,huang2025t2i}. Existing methods rely primarily on human annotations~\cite{otani2023toward}, while using automated metrics~\cite{jayasumana2024rethinking,yarom2023you,hessel2021clipscore}, preference-trained reward models, and vision--language model judges to assess image quality and human preferences~\cite{kirstain2023pick,liang2024rich,xu2024visionreward,wu2024multimodal}. Meanwhile, commercial recommendation research models preferences from implicit user feedback~\cite{he2016vbpr}, and visual commerce research has examined images in real-world commercial settings, including their relationships with social signals and art market value~\cite{ji2026servimage,lee2024social,ma2022colors}. These studies suggest that online commercial behavior provides observable feedback beyond human annotations; however, whether transaction prices can serve as reference signals for human visual preferences remains unverified.

As shown in Fig.~\ref{fig:compared-with-cheaptalk}(a), existing visual preference evaluation typically uses human annotations as a reference and employs preference-based reward models and vision--language model (VLM) judges to approximate human judgments~\cite{otani2023toward,kirstain2023pick,liang2024rich,xu2024visionreward,wu2024multimodal}; automated alignment metrics provide complementary signals for image evaluation~\cite{jayasumana2024rethinking,yarom2023you,hessel2021clipscore}. Meanwhile, related work combines implicit user feedback with visual features for personalized preference ranking~\cite{he2016vbpr}, or examines the relationships between visual and social attributes and art transaction prices~\cite{lee2024social,ma2022colors}, as shown in Fig.~\ref{fig:compared-with-cheaptalk}(b). However, using market information to support modeling or explain market outcomes does not establish that transaction prices represent human visual preferences. We therefore test whether transaction prices can serve directly as candidate labels for visual preference, as illustrated in Fig.~\ref{fig:compared-with-cheaptalk}(c). Specifically, we compare price-based rankings with price-blind human rankings to assess whether transaction prices are effective preference references.

To identify when transaction prices can serve as references for visual preference, we introduce \textbf{MarketImage}, a framework that separately diagnoses price signals, their consistency with human preferences, model recovery, and evaluation protocol effects. Specifically, \textbf{MarketImageArt} pairs realized transaction prices with price-blind human rankings across 266 six-image sets; \textbf{MarketImageLens} preserves price ties, distinguishes pairs where prices agree or conflict with human preferences, and audits display order; and \textbf{MarketValuator} evaluates static visual scorers and VLM judges on the conditionally consistent subset, comparing direct judgments with budget-constrained and multi-agent valuation interfaces. Of the 3,990 image pairs, only 24.2\% have a non-tied price direction, and just 34.7\% of those agree with the human majority. On the 335 protocol-supported pairs, HPSv2 reaches 66.4\% Recoverable Accuracy, while GPT-5.5 Pair and GPT-5.6 Sol Pair reach 60.3\% and 62.1\%, respectively. Qwen's budgeted and aggregated variants trail baselines; GPT interface gains lack statistical significance, while display-order audits reveal substantial confounding. These findings indicate that transaction prices provide a human-supported reference for visual preference only under limited conditions and should not be treated as ground truth without auditing.

The main contributions of this work are as follows: \textbf{MarketImageArt:} We construct a dataset of 266 six-image comparison sets, pairing realized transaction prices with price-blind human rankings over the same image sets. This provides an empirical basis for testing whether transaction prices reflect visual preferences.
\textbf{MarketImageLens:} We propose a tie-preserving pairwise diagnostic framework that distinguishes the availability of price directions, their consistency with human preferences, model recovery ability, and display-order effects. Our analysis shows that only 34.7\% of non-tied price directions agree with the human majority.
\textbf{MarketValuator:} We systematically evaluate the preference recovery ability of static visual scorers and multimodal language models, comparing direct judgments with budget-constrained and multi-agent evaluation interfaces. Qwen declines under budget and aggregation; GPT improvements remain statistically unresolved overall.

\section{Related Work}
\textbf{Visual Preference Evaluation and Market Signals.}
Existing visual evaluation methods use alignment metrics, aesthetic predictors, and preference or reward models trained from human feedback. CLIPScore~\cite{hessel2021clipscore} measures text--image alignment, while Pick-a-Pic~\cite{kirstain2023pick} supports learned preference scoring. Related studies examine self-supervised and multi-task aesthetic assessment, personalized aesthetic preferences, and art-preference prediction from visual features~\cite{sheng2020revisiting,soydaner2024multi,yang2022personalized,iigaya2021aesthetic}. Other work suggests that aesthetic judgments can be partly dissociated from semantic content~\cite{parraga2025aesthetics}, while recent reward models incorporate richer human feedback~\cite{liang2024rich,ba2025enhancing}. Preference-based VLM rewards have also been used to improve visual text rendering~\cite{cui2026textalign}. Fine-grained multimodal benchmarks diagnose visual-grounding failures through stage-wise metrics and expose VLM--human gaps on visual-text tasks~\cite{ji2026finestate,zhang2026revico}. Beyond explicit annotations, visual recommendation models use implicit user feedback to learn personalized rankings~\cite{he2016vbpr}, and visual-commerce research relates image properties and social signals to commercial outcomes, including human payment decisions and art-market value~\cite{ji2026servimage,lee2024social,ma2022colors}. However, whether observed transaction-price ordering can directly serve as a reference for human visual preference remains unclear. Unlike price prediction, our work examines the availability, human corroboration, and evaluator recoverability of price-implied preference directions.

\textbf{Market Mechanisms and Evaluation Uncertainty.}
Market mechanisms derive prices, allocations, or rankings through interactions involving agents, budgets, and competition. Prior studies investigate auction-based LLM agents~\cite{chen2023put}, fixed-budget voting for multi-issue decisions~\cite{georgescu2024fixed}, and token-curated registries for incentive-aware content curation~\cite{ito2019token}. Auction-inspired mechanisms have also been incorporated into generative modeling, including auction-driven diffusion and multi-player GAN training~\cite{sun2025bidding,shim2024bid}. Such interactions may produce strategic behavior, including collusion, that alters market outcomes~\cite{agrawal2025evaluating}; transaction prices may therefore contain visually relevant information while also reflecting market demand, competition, and allocation. Related LLM-based financial reporting workflows use evidence-grounded agents to normalize disclosures across jurisdictions~\cite{zhang2026finreporting}; that setting differs from using realized market prices as candidate visual-preference signals. Complementary benchmarks evaluate multi-agent debate across domains and modalities~\cite{li2026m3madbench}. Human and model judgments can also vary with evaluation protocols, including grading scales and stylistic presentation~\cite{otani2023toward,li2026grading,ji2026stylewins}. Motivated by these uncertainties, we separately diagnose price-reference ambiguity, presentation-order confounds, model recovery, and elicitation effects rather than interpreting aggregate agreement as visual-preference accuracy.

\begin{figure*}[t]
\centering
\includegraphics[width=\textwidth]{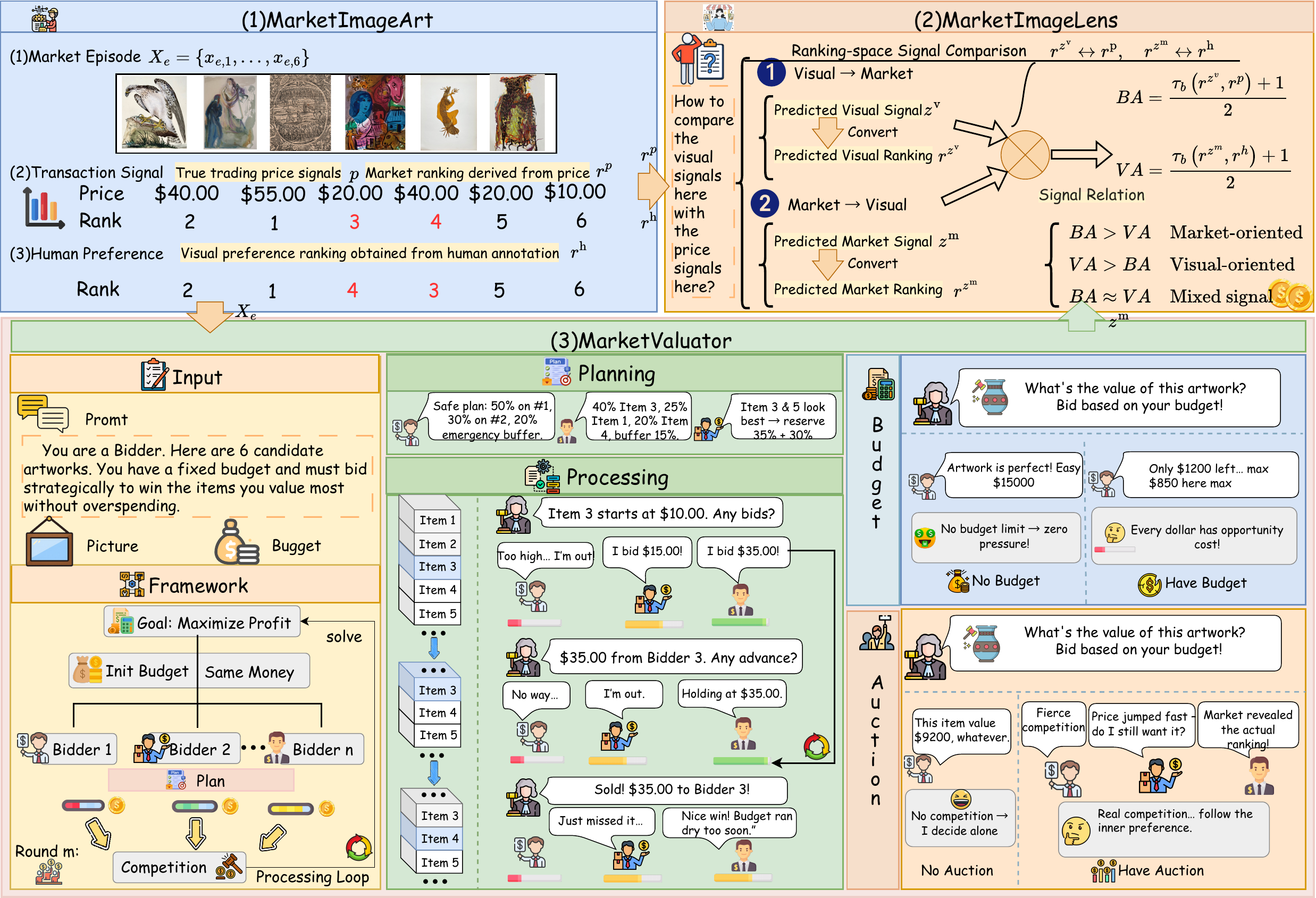}
\caption{
Overview of the MarketImage workflow. (1) MarketImageArt pairs each
constructed six-image set with transaction prices and price-blind human
rankings; sets are not claimed to be single observed auctions. (2)
MarketImageLens compares visual-evaluator rankings with price rankings and
MarketValuator rankings with human rankings, preserving ties. (3) MarketValuator
illustrates image-based, multi-agent elicitation with and without budget and
auction constraints; the evaluated interfaces do not replay the original
transactions. Here, $r^p$, $r^h$, $r^{z^v}$, and $r^{z^m}$ denote price, human,
visual-evaluator, and MarketValuator rankings; BA and VA are normalized rank
agreements for the two comparisons in (2). Source order is audited separately
as a protocol confound.
}
\label{fig:overall}
\end{figure*}

\section{MarketImageArt}
\subsection{Task Formulation}
For comparison set $e$, let
$X_e=\{x_{e,1},\ldots,x_{e,n}\}$ denote its candidate images, where
$e\in\{1,\ldots,E\}$ and $n=6$. These are benchmark evaluation sets formed
within a price stratum; they are not claimed to be six lots from one observed
auction. Panel~(1) of Figure~\ref{fig:overall} illustrates one such set
alongside its transaction-price signal and human-preference ranking. Artwork
$x_{e,i}$ has transaction price $p_{e,i}$ and unobserved
non-visual market factors $\mathbf m_{e,i}$, such as provenance, timing,
demand, and competition. We represent price as
\begin{equation}
p_{e,i}=\mathcal G(\mathbf v_{e,i},\mathbf m_{e,i}),
\label{eq:mixed_price}
\end{equation}
where $\mathbf v_{e,i}$ denotes latent visually grounded factors and
$\mathcal G$ is an unknown price-formation process. This is a conceptual
decomposition: we do not estimate $\mathcal G$ or causally separate its
inputs. Instead, we audit whether pairwise directions derived from the
observed vector $\mathbf p_e=(p_{e,1},\ldots,p_{e,n})$ are available and
corroborated by independently elicited visual judgments.

\subsection{Benchmark Construction}
MarketImageArt is constructed from real emerging-art auction records through
data cleaning and comparison-set organization.
\textbf{Data cleaning:}
We initially restrict prices to EUR~10--1,000 and retain art-print listings
with valid normalized prices and retrievable images. We remove inconsistent
records and evaluator-visible auction metadata. The final sample contains
1,596 artworks priced from EUR~10 to EUR~395 over 2020--2024.

\begin{table}[t]
\centering
\small
\setlength{\tabcolsep}{3pt}
\renewcommand{\arraystretch}{0.4}
\caption{MarketImageArt price statistics. Low-, mid-, and high-price sets partition the 1,596 items; Near-tie-low and Near-tie-high are derived 20\% set views, not additional artworks.}
\label{tab:marketimageart_price_buckets}
\resizebox{\columnwidth}{!}{%
\begin{tabular}{lrrrrrr}
\toprule
\multicolumn{7}{c}{\textbf{Overall Price Statistics (EUR)}} \\
\midrule
\textbf{Subset} & \textbf{Min} & \textbf{Q25} & \textbf{Median} & \textbf{Mean} & \textbf{Q75} & \textbf{Max} \\
\midrule
Overall & 10 & 40 & 70 & 96.9 & 121 & 395 \\
\midrule
\multicolumn{7}{c}{\textbf{Normal Scenarios}} \\
\midrule
\textbf{Subset} & \textbf{\#Items} & \textbf{\#Groups} & \textbf{Min} & \textbf{Max} & \textbf{Mean} & \textbf{Median} \\
\midrule
Overall    & 1596 & 266 & 10  & 395 & 96.9  & 70  \\
Low-price  & 858  & 143 & 10  & 75  & 43.2  & 45  \\
Mid-price  & 504  & 84  & 75  & 170 & 112.2 & 110 \\
High-price & 234  & 39  & 170 & 395 & 260.5 & 250 \\
\midrule
\multicolumn{7}{c}{\textbf{Stress-Test Scenarios}} \\
\midrule
Near-tie-low  & 318 & 53 & 10 & 45  & 30.1  & 30 \\
Near-tie-high & 318 & 53 & 12 & 390 & 121.5 & 78 \\
\bottomrule
\end{tabular}%
}
\end{table}

\textbf{Comparison-set construction:}
As shown in Table~\ref{tab:marketimageart_price_buckets}, the artworks are
partitioned into low-, mid-, and high-price strata and organized into 266
six-image sets. Low-, mid-, and high-price subsets are disjoint views of these
sets. Near-tie-low and Near-tie-high each select 53 sets from Overall by the
smallest and largest within-set relative price gap. They are derived stress
tests rather than independent samples. A post-construction audit found that
every canonical set is stored in non-decreasing price order; we retain this
historical order, disclose it, and test model-side order sensitivity instead of
silently rewriting completed experiments.

\textbf{Visual preference annotation:}
The released reference uses three of four complete annotator files; the
exclusion rationale is undocumented. We use this three-annotator consensus
for the primary analysis and report the all-four sensitivity analysis in
Appendix~\ref{sec:appendix} (Table~\ref{tab:appendix_annotator_sensitivity}).
Each included annotator completed all 266 Overall sets. Annotators saw the six
images, folder IDs, and numeric source-position labels, but not prices,
estimates, or artwork metadata. Images remained in canonical source order, so
the annotations are price-blind but not order-randomized. The primary human
reference is the annotator-averaged pairwise preference probability; a Kemeny
consensus order is retained only for visualization. Pairwise annotator
agreements are \(65.8\%\), \(61.8\%\), and \(67.6\%\) (Kendall's \(\tau=0.316\),
\(0.236\), and \(0.352\)); leave-one-out agreement with the other two
annotators' pairwise consensus is \(70.4\%\), \(77.0\%\), and \(72.3\%\).
\textbf{Reference-signal construction:}
MarketImageArt thus provides aligned transaction-price and price-blind
human-preference signals. Section~\ref{subsec:lens} defines how MarketImageLens
compares them pairwise while preserving price ties.

\subsection{Benchmark Validation}
\label{subsec:benchmark_validation}
Appendix Figure~\ref{fig:emergingart_validation} summarizes the data-quality
and set-structure audit. The sample spans EUR~10--395 (median EUR~70) over
2020--2024; the image payload excludes explicit prices, estimates, lot
identifiers, URLs, and auction metadata. Only 2 of 1,596 titles retain an
artist-name mention, and titles are not supplied to evaluated models. The 266
six-image sets are item-disjoint and concentrated in an outcome-near-tie
regime: 87.2\% have zero top-two price gap and 57.9\% have all six prices
tied. Canonical order is price-sorted, making source position an implicit
target cue. Appendix~\ref{app:reference_order} reports the full reference audit
and seeded model-side order intervention. Thus, explicit metadata leakage is
controlled, but presentation-order exposure remains part of the benchmark.

\section{MarketImage Framework}
\label{sec:method}
Figure~\ref{fig:overall} summarizes the three-stage workflow: MarketImageArt
provides aligned transaction-price and human-preference signals, MarketImageLens
audits their pairwise relation and evaluator recovery, and MarketValuator tests
whether alternative elicitation interfaces change that recovery. The framework
keeps reference validity, evaluator performance, and interface effects
analytically separate.

\subsection{MarketImageLens}
\label{subsec:lens}
MarketImageLens analyzes heterogeneous value signals in a common pairwise
space while preserving ties. For a set-level signal \(\mathbf{s}_e\),
the direction of a within-set comparison is
\begin{equation}
d^{s}_{e,ij}=\operatorname{sign}(s_{e,i}-s_{e,j}),
\quad
\mathbf{s}_e\in\{\mathbf{p}_e,\mathbf{z}^{v}_e,\mathbf{z}^{m}_e\},
\label{eq:pairwise_space}
\end{equation}
where \(\mathbf{p}_e\) is transaction price, \(\mathbf{z}^{v}_e\) is a
visual-evaluator output, and \(\mathbf{z}^{m}_e\) is a signal elicited by
MarketValuator. The human direction \(d^h_{e,ij}\) is obtained from the
price-blind pairwise majority under the recorded fixed presentation order.
Because the human reference preserves that source order, these comparisons are
protocol-conditional; Appendix~\ref{app:reference_order} details the reference
audit and seeded order intervention.

The lens partitions comparisons into three observable reference regimes:
price--human agreement, price--human conflict, and price ties. It first audits
whether price provides a direction and whether that direction is corroborated
by the recorded human majority. Evaluator recovery is then assessed on the
protocol-supported subset, while conflict and tied-price pairs remain separate
diagnostic regions. Finally, paired changes between direct, budgeted, and
multi-profile interfaces measure system-level elicitation contrasts within
a fixed base model.
This decomposition prevents a model error, an uninformative price tie, and a
visually unsupported market outcome from receiving the same interpretation.
Table~\ref{tab:failure_map} summarizes the diagnostics and conclusions licensed
at each stage.

\begin{table*}[t]
\centering
\small
\renewcommand{\arraystretch}{0.6}
\setlength{\tabcolsep}{4pt}
\caption{Stage-wise decision map. $P$, $H$, and $\widehat{Y}$ denote price, human-majority, and evaluator pair directions; downstream scores cannot validate an unsupported reference.}
\label{tab:failure_map}
\begin{tabular}{llll}
\toprule
\textbf{Stage} & \textbf{Observable condition} & \textbf{Primary diagnostic} & \textbf{Licensed conclusion} \\
\midrule
Reference & $P=0$ or $P\ne H$ & Coverage; price--human agreement & Price direction is absent or lacks human support \\
Protocol & Position correlates with $P$ or $H$ & Source-order audit; permutation & Reference or prediction is order-confounded \\
Recovery & $P=H\ne0$, but $\widehat{Y}\ne P$ & Conditional accuracy; ties and margin & Evaluator disagrees or ties on a supported reference \\
Elicitation & Mechanism changes direct baseline & Paired contrast; correction/harm & Interface repairs, preserves, or damages judgments \\
\bottomrule
\end{tabular}
\end{table*}

\subsection{MarketValuator}
\label{subsec:valuator}

\begin{figure}[t]
\centering
\includegraphics[width=\columnwidth]{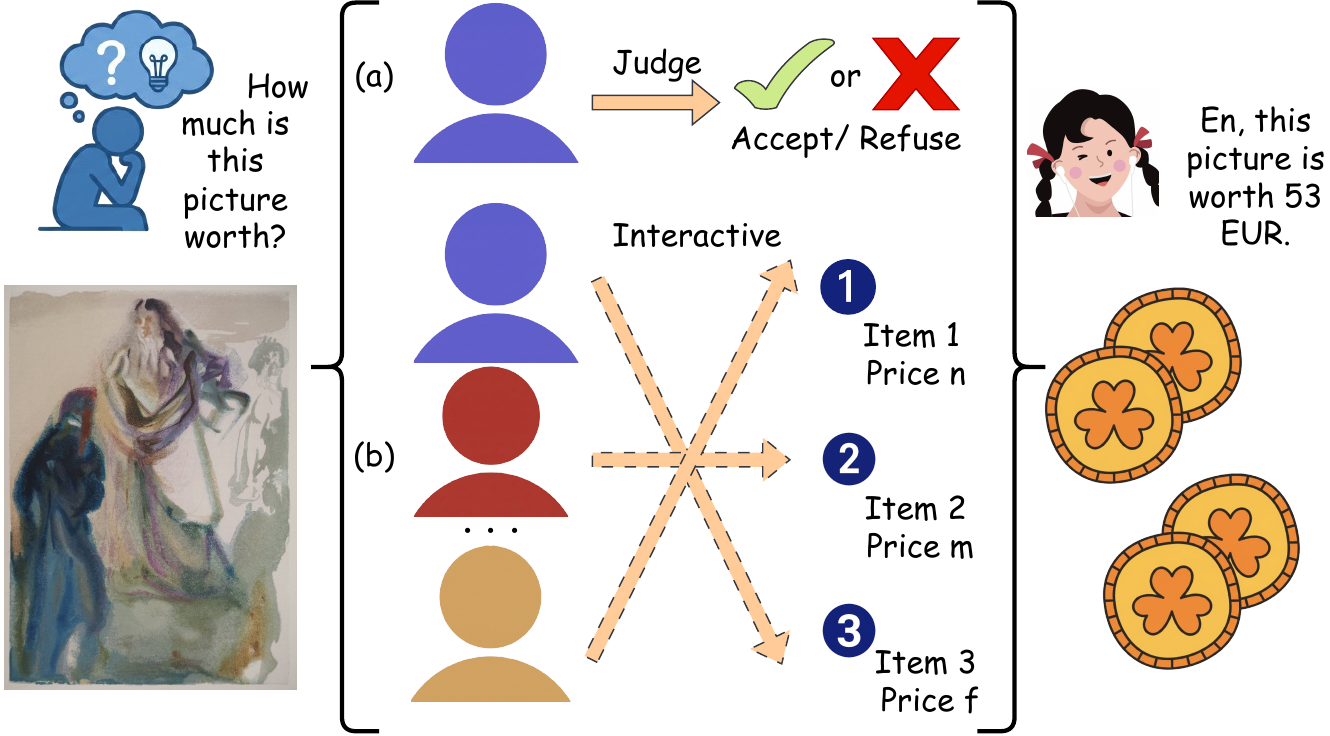}
\caption{Conceptual comparison between direct visual judgment and
budget-constrained multi-agent valuation. The implementations evaluated in
Section~\ref{subsec:valuator} are elicitation interfaces, not reproductions of
the original auctions.}
\Description{Diagram contrasting direct visual judgment with budget-constrained multi-agent valuation.}
\label{fig:marketvaluator_motivation}
\end{figure}

As illustrated in Figure~\ref{fig:marketvaluator_motivation}, MarketValuator
varies how image-based relative values are elicited; the tested interfaces do
not replay the original auctions. Given comparison set $X_e$, bidder profile
$a$, and budget $B_a$, the simultaneous interface returns
\begin{equation}
\begin{aligned}
\mathbf b_{a,e}&=\phi_a(X_e,B_a),\\
b_{a,e,i}&\geq0,\qquad \sum_{i=1}^{n}b_{a,e,i}\leq B_a.
\end{aligned}
\label{eq:budgeted_bid}
\end{equation}
All candidates are therefore judged jointly under a shared resource constraint.
With three bidder profiles, the reported API-native Valuator signal is the
per-item second-highest bid,
\begin{equation}
z^m_{e,i}=\operatorname{2ndmax}_{a\in\{1,2,3\}} b_{a,e,i}.
\label{eq:simultaneous_clear}
\end{equation}
Valuator Judge is the non-monetary ablation: each profile returns a 0--100
vector and $z^m_{e,i}$ is their mean. Budgeted Single-agent uses one constrained
vector without profile aggregation.

The legacy Qwen Valuator follows a different call graph. It presents items
sequentially, retains bidding history and remaining-budget context, and uses
the terminal accepted bid as its score. The API-native replication instead
elicits three independent simultaneous allocations and applies
Equation~\ref{eq:simultaneous_clear}; the bidders do not observe one another's
responses. We preserve both completed implementations but compare an interface
only with baselines using the same base model. These are system-level contrasts,
not token-matched estimates of a pure competition effect. MarketImageLens then
maps every resulting vector into the common tie-preserving pairwise space.
Appendix~\ref{app:mechanism_diagnostics} reports the complementary audit of
budget use, bidder disagreement, and aggregation behavior.

\section{Experiments}
\subsection{Diagnostic Questions}
The experiments do not assume that transaction prices are visual-preference
labels. Instead, they separate three questions that were conflated by the
set-level evaluation in the earlier design:
\textbf{RQ1} asks when transaction-price order is identifiable and agrees with
price-blind human visual preference under the observed presentation protocol;
\textbf{RQ2} asks whether an evaluator recovers price order on that
protocol-supported subset; and
\textbf{RQ3} asks whether budgeted and multi-profile interfaces improve or
damage this recovery relative to matched comparison methods. This ordering is important: mechanism
performance cannot establish label validity, and agreement with a price tie
cannot establish directional recovery.

\subsection{Experimental Setup}
\label{subsec:experimental_setup}
\paragraph{Evaluators.}
We re-score the eight static visual evaluators used in the original experiment
(CLIPScore, NIMA, HPSv2, PickScore, ImageReward, RichHF, Q-Align, and
LLaVA-Reward) and six Qwen3-VL-4B elicitation variants: Single-agent (Pair),
Single-agent, Single-agent (LLM Score), Budgeted Single-agent, Valuator Judge,
and Valuator. We additionally replicate all six interfaces with GPT-5.5 and
GPT-5.6 Sol (runner identifiers \texttt{gpt-5.5} and
\texttt{gpt-5.6-sol}) through ephemeral Codex-hosted sessions with medium
reasoning effort. Every row covers the same 266 comparison sets. Joint interfaces use the
price-sorted canonical source order; outputs are aligned by the unordered set
of internal item identifiers and scored in canonical orientation.
Appendix~\ref{app:reference_order} audits this exposure, and
Appendix~\ref{app:order_intervention} reruns the Pair interface under seeded
permutations for the two GPT models. Higher scores indicate stronger preference
or value.

\paragraph{API interface replication.}
The API runs preserve the information and decision constraints of each
interface while using structured outputs. Single-agent independently predicts
one price per image, and Single-agent (LLM Score) independently assigns one
0--100 score per image. Single-agent (Pair) ranks all six images jointly.
Budgeted Single-agent allocates bids under a EUR~600 set-level budget cap.
Valuator Judge averages scores from three independent judge profiles, whereas
Valuator uses three independently budgeted bidder profiles, each with the same
cap, and the per-item second-highest bid as its aggregation statistic. The
controlled user instruction contains no transaction price, human preference,
or artwork metadata and explicitly prohibits using file names. Images are
attached by the runner, and joint calls retain canonical position. These runs
are API-native replications of the six elicitation interfaces rather than
token-identical replays of the local Qwen implementation; raw outputs and
allocation traces are retained for audit.

\paragraph{Canonical pair index.}
Each six-image set contributes $\binom{6}{2}=15$ within-set pairs,
giving
\begin{equation}
\begin{aligned}
\mathcal U&=\{(e,i,j):e=1,\ldots,266,\ i<j\},\\
|\mathcal U|&=3{,}990.
\end{aligned}
\end{equation}
We never compare images across sets because human references are collected
within-set and joint budgeted outputs are not calibrated across sets. For pair
$k=(e,i,j)$, the price direction is
\begin{equation}
P_k=\operatorname{sign}(p_{e,i}-p_{e,j})\in\{-1,0,1\},
\end{equation}
where $P_k=0$ denotes an exact transaction-price tie. From the three complete
human rankings, we compute
\begin{align}
q_k&=\frac{1}{3}\sum_{a=1}^{3}\mathbb{I}[r_a(i)<r_a(j)],\\
H_k&=\operatorname{sign}(q_k-0.5).
\end{align}
Thus $q_k\in\{0,1/3,2/3,1\}$ and $H_k$ has no tie. We retain both 2:1 majority
and 3:0 unanimous judgments and report them separately for RQ1.

\paragraph{Reference reliability (RQ1).}
Price Coverage is the fraction of pairs for which $P_k\ne0$. Let
$N_P=|\{k:P_k\ne0\}|$ and $a_k=q_k$ if $P_k=1$, otherwise
$a_k=1-q_k$. Conditional on price-ordered pairs, hard and soft price--human
agreement are
\begin{align}
A_{PH}&=\frac{1}{N_P}\sum_{k:P_k\ne0}\mathbb{I}[P_k=H_k],\\
A_{PH}^{\mathrm{soft}}&=\frac{1}{N_P}\sum_{k:P_k\ne0}a_k.
\end{align}
Price-tied pairs are excluded from both denominators rather than being treated
as chance-level agreement.

\paragraph{Evaluator recovery (RQ2).}
For evaluator scores $z$, the predicted direction is
$\widehat{Y}_k=\operatorname{sign}(z_{e,i}-z_{e,j})$. We preserve exact
prediction ties and never jitter scores or randomly break rankings. The primary
recovery set is
\begin{equation}
\mathcal C=\{k:P_k=H_k\ne0\},
\end{equation}
which contains pairs for which the observed price direction agrees with the
price-blind human majority under the recorded presentation protocol. We call
this subset \emph{protocol-supported}; membership is observational and does not
identify a causal visual component of price. Define the directional score
\begin{equation}
\ell(\widehat{Y}_k,T_k)=\mathbb{I}[\widehat{Y}_k=T_k]
+\tfrac{1}{2}\mathbb{I}[\widehat{Y}_k=0],\quad T_k\in\{-1,1\}.
\end{equation}
Recoverable Accuracy is
\begin{equation}
\operatorname{RA}=\frac{1}{|\mathcal C|}\sum_{k\in\mathcal C}
\ell(\widehat{Y}_k,P_k),
\end{equation}
where exact prediction ties receive half credit, and we report their frequency
separately to expose methods that collapse to ties. For any pair subset $S$ and
target vector $T=(T_k)_{k\in S}$, define
$\operatorname{Acc}(S,T)=|S|^{-1}\sum_{k\in S}\ell(\widehat{Y}_k,T_k)$.
Human Accuracy is $\operatorname{Acc}(\mathcal U,H)$; Price Accuracy is
$\operatorname{Acc}(\{k:P_k\ne0\},P)$; Conflict$\rightarrow$H is
$\operatorname{Acc}(\{k:P_k\ne0, P_k\ne H_k\},H)$; and Price-tie$\rightarrow$H
is $\operatorname{Acc}(\{k:P_k=0\},H)$. Thus, the latter two columns measure
each evaluator's agreement with the human majority within those reference
strata, not inter-annotator agreement.

\paragraph{Uncertainty estimates and completeness.}
All confidence intervals use 10,000 percentile-bootstrap samples over the 266
sets, not independent pair resampling. This preserves dependence among the
15 pairs in a set. A method is included in the main table only when all
266 canonical sets align successfully. The API-native runs additionally retain
parse, fallback, and trace status. Historical static and Qwen files that expose
only final scores are treated as aligned legacy artifacts, not as proof that no
internal fallback occurred.

\begin{table}[t]
\centering
\renewcommand{\arraystretch}{0.6}
\setlength{\tabcolsep}{4.0pt}
\caption{Reference decomposition over all 3,990 within-set pairs. The 2:1 and 3:0 columns denote majority and unanimous human judgments. Price ties are not assigned a price direction.}
\label{tab:reference_decomposition}
\resizebox{\columnwidth}{!}{%
\begin{tabular}{lrrr}
\toprule
\textbf{Pair class} & \textbf{2:1} & \textbf{3:0} & \textbf{Total} \\
\midrule
Price--human agree & 228 & 107 & 335 \\
Price--human conflict & 298 & 333 & 631 \\
Price tie & 1565 & 1459 & 3024 \\
\midrule
Total & 2091 & 1899 & 3990 \\
\bottomrule
\end{tabular}
}
\end{table}

\subsection{RQ1: When Is Price Direction Identifiable and Human-Corroborated?}
To determine whether transaction prices provide a usable visual-preference
reference before evaluating models, we measure price-direction coverage and
agreement with price-blind human judgments over the 3,990 within-set pairs
(definitions above; Table~\ref{tab:reference_decomposition}). Price direction
is sparse and weakly corroborated: only 966 pairs have a non-tied price order
(24.2\% coverage), with human agreement on 335 and conflict on 631. This gives
34.7\% hard agreement (95\% set-cluster CI: 30.2--39.0) and 37.1\% soft
agreement.

Clearer human consensus does not yield stronger price agreement: it is 24.3\%
for unanimous pairs versus 43.3\% for 2:1 majorities. Further, every non-tied
price direction favors the later canonical item, while the human majority
favors the earlier item in 65.9\% of pairs (Table~\ref{tab:human_order}); all
three annotators show this positional tendency. Thus, the observed agreement
mixes reference disagreement with fixed-order exposure and cannot identify the
visual contribution to price. We do not treat raw transaction order as visual
ground truth and use the 335 concordant pairs only for conditional recovery.

\begin{table}[t]
\centering
\renewcommand{\arraystretch}{0.4}
\setlength{\tabcolsep}{5pt}
\caption{Human preference by initial source position over all 3,990 pairs.}
\label{tab:human_order}
\begin{tabular}{lrr}
\toprule
\textbf{Preference source} & \textbf{Earlier (\%)} & \textbf{Later (\%)} \\
\midrule
Annotator A & 56.9 & 43.1 \\
Annotator B & 63.1 & 36.9 \\
Annotator C & 70.8 & 29.2 \\
\midrule
Human majority & 65.9 & 34.1 \\
\bottomrule
\end{tabular}
\end{table}

\subsection{RQ2: Can Models Recover the Protocol-Supported Price Order?}
\begin{table*}[t]
\centering
\small
\renewcommand{\arraystretch}{0.6}
\setlength{\tabcolsep}{3.0pt}
\caption{Pairwise diagnosis on MarketImageArt under the recorded fixed presentation protocol. Rec. Acc. is evaluated only on the 335 pairs for which transaction-price order and the human majority agree. Human Acc. uses all 3,990 pairs; Price Acc. uses the 966 non-price-tied pairs. Conflict$\rightarrow$H and Price-tie$\rightarrow$H report evaluator agreement with the human majority on the 631 conflicting and 3,024 tied-price pairs, respectively. Exact prediction ties receive half credit in directional accuracies; Pred. tie reports their frequency over all 3,990 pairs. Confidence intervals use 10,000 set-cluster bootstrap samples.}
\label{tab:pairwise_diagnosis}
\resizebox{\textwidth}{!}{%
\begin{tabular}{lrrrrrrr}
\toprule
\textbf{Method} & \textbf{Sets/Rec.} & \textbf{Rec. Acc. [95\% CI]} & \textbf{Human Acc.} & \textbf{Price Acc.} & \textbf{Conflict$\rightarrow$H} & \textbf{Price-tie$\rightarrow$H} & \textbf{Pred. tie} \\
\midrule
\multicolumn{8}{c}{\textit{Static visual preference models}} \\
CLIPScore & 266/335 & 46.0 [38.6, 53.3] & 48.1 & 47.8 & 51.2 & 47.7 & 0.4 \\
NIMA & 266/335 & 63.6 [56.9, 70.2] & 56.6 & 51.4 & 55.0 & 56.2 & 0.0 \\
HPSv2 & 266/335 & 66.4 [59.1, 73.2] & 61.8 & 49.5 & 59.5 & 61.8 & 0.3 \\
PickScore & 266/335 & 66.0 [60.0, 72.2] & 59.8 & 47.0 & 63.1 & 58.4 & 0.0 \\
ImageReward & 266/335 & 63.6 [56.9, 70.0] & 59.8 & 49.2 & 58.5 & 59.7 & 0.0 \\
RichHF & 266/335 & 40.0 [33.7, 46.5] & 43.6 & 50.6 & 43.7 & 44.0 & 0.0 \\
Q-Align & 266/335 & 43.6 [36.4, 50.8] & 47.2 & 46.9 & 51.3 & 46.7 & 0.2 \\
LLaVA-Reward & 266/335 & 49.7 [43.4, 56.2] & 48.6 & 49.1 & 51.3 & 47.9 & 0.9 \\
\midrule
\multicolumn{8}{c}{\textit{Qwen3-VL-4B elicitation methods}} \\
Single-agent (Pair) & 266/335 & 8.4 [4.6, 13.0] & 59.5 & 12.1 & 85.9 & 59.7 & 0.0 \\
Single-agent & 266/335 & 51.5 [46.5, 56.4] & 51.0 & 50.2 & 50.5 & 51.0 & 61.2 \\
Single-agent (LLM Score) & 266/335 & 42.7 [36.1, 49.7] & 47.7 & 47.4 & 50.2 & 47.7 & 19.1 \\
Budgeted Single-agent & 266/335 & 41.5 [33.1, 50.0] & 51.6 & 43.6 & 55.3 & 51.9 & 0.1 \\
Valuator Judge & 266/335 & 44.3 [37.7, 50.9] & 49.9 & 44.9 & 54.8 & 49.5 & 17.3 \\
Valuator & 266/335 & 24.9 [20.0, 29.9] & 55.2 & 27.8 & 70.6 & 55.3 & 5.4 \\
\midrule
\multicolumn{8}{c}{\textit{GPT-5.5 elicitation methods}} \\
Single-agent (Pair) & 266/335 & 60.3 [52.0, 68.2] & 55.6 & 50.5 & 54.7 & 55.2 & 0.0 \\
Single-agent & 266/335 & 45.8 [38.8, 53.1] & 47.2 & 51.4 & 45.6 & 47.7 & 8.3 \\
Single-agent (LLM Score) & 266/335 & 55.8 [47.9, 63.6] & 57.0 & 46.8 & 57.9 & 57.0 & 12.8 \\
Budgeted Single-agent & 266/335 & 48.8 [40.3, 57.4] & 55.2 & 42.9 & 60.3 & 54.9 & 0.3 \\
Valuator Judge & 266/335 & 52.7 [45.1, 60.0] & 55.3 & 44.5 & 59.8 & 54.7 & 0.6 \\
Valuator & 266/335 & 51.9 [43.7, 60.1] & 51.9 & 48.3 & 53.6 & 51.6 & 1.5 \\
\midrule
\multicolumn{8}{c}{\textit{GPT-5.6 Sol elicitation methods}} \\
Single-agent (Pair) & 266/335 & 62.1 [54.0, 69.9] & 58.9 & 48.9 & 58.2 & 58.8 & 0.0 \\
Single-agent & 266/335 & 52.8 [46.1, 59.4] & 47.3 & 57.2 & 40.5 & 48.1 & 7.9 \\
Single-agent (LLM Score) & 266/335 & 60.4 [52.8, 67.7] & 57.9 & 49.3 & 56.6 & 57.9 & 8.5 \\
Budgeted Single-agent & 266/335 & 58.8 [50.6, 66.7] & 58.2 & 46.1 & 60.7 & 57.6 & 0.1 \\
Valuator Judge & 266/335 & 55.2 [47.1, 62.9] & 56.3 & 45.9 & 59.1 & 55.8 & 0.8 \\
Valuator & 266/335 & 57.6 [50.0, 65.0] & 56.2 & 48.1 & 56.9 & 55.9 & 1.5 \\
\bottomrule
\end{tabular}%
}
\end{table*}

To test whether evaluators recover price directions corroborated by humans, we
score all static models and elicitation interfaces on the same 335-pair
protocol-supported set. Table~\ref{tab:pairwise_diagnosis} also reports
complementary measures on the broader reference strata. Overall, recovery is
partial and varies across systems; no method establishes transaction price as
general visual ground truth.

Among static scorers, HPSv2 leads at 66.4\% (95\% CI: 59.1--73.2), followed
by PickScore at 66.0\%; NIMA and ImageReward each reach 63.6\%, while RichHF
is lowest at 40.0\%. The leading static models therefore recover part of the
price--human relation, but the selected subset retains the order confound.
Across model families, GPT-5.5 and GPT-5.6 Sol Pair reach 60.3\% and 62.1\%,
respectively, versus 8.4\% for Qwen3-VL-4B Pair; their 0--100 visual-score
interfaces also exceed the corresponding Qwen interface (55.8\% and 60.4\%
vs. 42.7\%). This is not uniform across interfaces: GPT-5.5 Single-agent
scores 45.8\%, below Qwen3-VL-4B Single-agent at 51.5\%.

The low Qwen Pair recovery is especially notable: it has 59.5\% Human Accuracy
and 85.9\% agreement on price--human conflicts, yet reaches only 8.4\% and
favors the earlier canonical item in 83.3\% of decisive comparisons
(Appendix~\ref{app:reference_order}), consistent with order sensitivity.
Qwen Single-agent's 51.5\% also coincides with a 61.2\% prediction-tie rate,
so it does not indicate reliable directional recovery. GPT-5.6 Sol lowers the
direct price predictor's tie rate to 7.9\%. HPSv2 nevertheless remains the
highest-scoring method, with a confidence interval overlapping both GPT Pair
interfaces; the observed differences do not establish that agent elicitation
outperforms static visual scoring.

\subsection{RQ3: Do Budgeted and Multi-Profile Interfaces Help?}
To test whether budget constraints and profile aggregation improve recovery,
we compare interfaces within each model family on the same 335
protocol-supported pairs. Table~\ref{tab:elicitation_contrasts} reports paired
Recoverable Accuracy change $\Delta=\operatorname{RA}_{\mathcal C}(E_{\rm variant})-
\operatorname{RA}_{\mathcal C}(E_{\rm comparator})$, where each contrast row
specifies the tested interface and comparator, along with correction and harm rates and
set-cluster bootstrap intervals. Correction counts baseline errors fixed
exactly by the tested interface; Harm counts baseline exact-correct predictions
made non-correct. Ties receive half credit for accuracy but count as non-correct
for these two rates.

\begin{table}[H]
\centering
\scriptsize
\renewcommand{\arraystretch}{0.4}
\setlength{\tabcolsep}{2.5pt}
\caption{Paired contrasts on the 335 protocol-supported pairs. $\Delta$ and confidence intervals are percentage-point changes. Corr. is the percentage of baseline non-correct decisions corrected exactly; Harm is the percentage of baseline exact-correct decisions made non-correct. Ties count as non-correct for these two rates. Intervals are 95\% set-cluster bootstrap CIs.}
\label{tab:elicitation_contrasts}
\resizebox{\columnwidth}{!}{%
\begin{tabular}{lrrr}
\toprule
\textbf{Contrast} & $\Delta$ \textbf{Rec. Acc. [95\% CI]} & \textbf{Corr.} & \textbf{Harm} \\
\midrule
\multicolumn{4}{l}{\textit{Qwen3-VL-4B}} \\
Budgeted $-$ Single & -10.0 [-18.9, -1.0] & 37.7 & 45.3 \\
Judge $-$ Single & -7.2 [-15.4, 1.2] & 33.5 & 64.0 \\
Valuator $-$ Judge & -19.4 [-25.3, -13.7] & 6.3 & 49.1 \\
Valuator $-$ Single & -26.6 [-32.7, -20.5] & 20.4 & 74.7 \\
\midrule
\multicolumn{4}{l}{\textit{GPT-5.5}} \\
Budgeted $-$ Single & 3.0 [-6.7, 12.2] & 37.9 & 36.4 \\
Judge $-$ Single & 6.9 [-1.5, 15.0] & 43.1 & 34.3 \\
Valuator $-$ Judge & -0.7 [-4.5, 2.9] & 11.3 & 13.6 \\
Valuator $-$ Single & 6.1 [-3.2, 15.3] & 39.5 & 33.6 \\
\midrule
\multicolumn{4}{l}{\textit{GPT-5.6 Sol}} \\
Budgeted $-$ Single & 6.0 [-3.4, 14.9] & 48.2 & 30.8 \\
Judge $-$ Single & 2.4 [-5.9, 10.4] & 40.4 & 31.4 \\
Valuator $-$ Judge & 2.4 [-1.4, 6.3] & 15.1 & 9.8 \\
Valuator $-$ Single & 4.8 [-3.5, 13.0] & 44.0 & 32.0 \\
\bottomrule
\end{tabular}%
}
\end{table}

\begin{figure*}[t]
\centering
\scalebox{1}[0.95]{%
\includegraphics[width=0.82\textwidth]{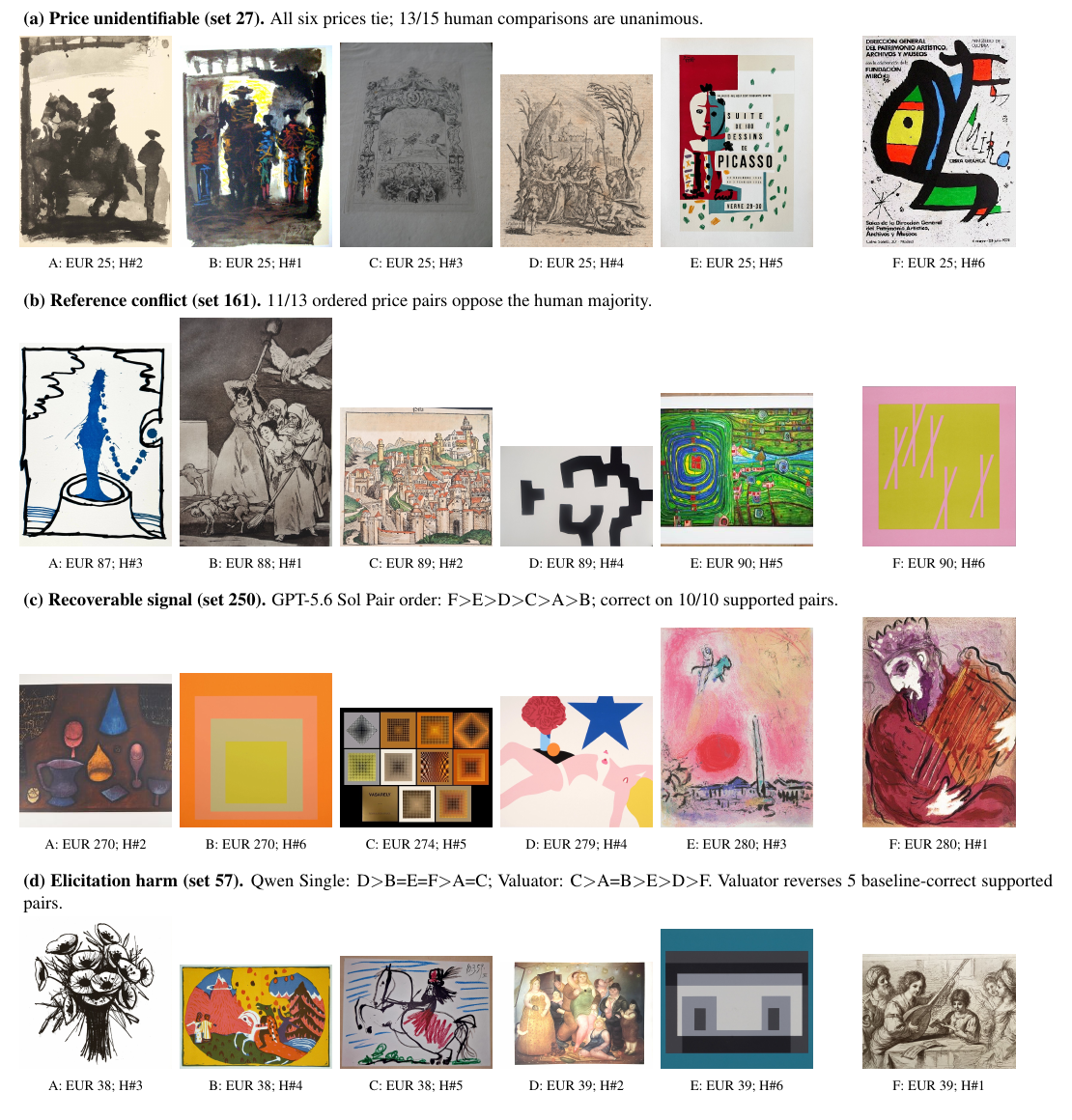}%
}
\caption{Qualitative audit of the four-stage diagnosis. A--F denote canonical
item identity, EUR is observed transaction price, and H\# is Kemeny human rank
(1 is best). Cases are selected deterministically: (a) maximizes unanimous
human comparisons among all-price-tie sets; (b) maximizes unanimous
price--human conflicts; (c) maximizes correctly recovered supported pairs for
GPT-5.6 Sol Pair; and (d) maximizes Qwen Valuator harm minus correction relative
to Single-agent. These diagnostic extremes are not representative performance
estimates.}
\label{fig:qualitative_cases}
\end{figure*}

The paired results show no general recovery gain from adding budget or
multi-profile aggregation. The clearest negative effects occur for Qwen3-VL-4B: Budgeted
Single-agent is 10.0 points below Single-agent (95\% CI: $-18.9$ to $-1.0$),
while Valuator is 19.4 points below Valuator Judge and 26.6 points below
Single-agent. Valuator corrects 20.4\% of Single-agent's non-correct decisions
but harms 74.7\% of its exact-correct decisions, making it the most harmful
tested contrast in this family.

The pattern differs for the GPT systems. All GPT confidence intervals include
zero: changes relative to Single-agent range from +3.0 to +6.9 points for
GPT-5.5 and +2.4 to +6.0 points for GPT-5.6 Sol, with the largest increase
for GPT-5.5 Valuator Judge. Valuator versus Valuator Judge changes recovery by
$-0.7$ and +2.4 points, respectively. Thus, these positive point estimates do
not resolve into a statistically supported advantage, unlike the measured
Qwen degradation. These are
system-level contrasts rather than isolated competition effects because the
interfaces are not compute-matched. All 3,192 API method--set outputs align
without parse failure or fallback. Figure~\ref{fig:qualitative_cases}
illustrates selected examples of the distinct failure modes; they are
diagnostic cases, not prevalence estimates. Additional appendix results assess
annotator-count and relative-price-gap sensitivity
(Tables~\ref{tab:appendix_annotator_sensitivity} and
\ref{tab:appendix_relative_gap}), alternative tie scoring and margin-based
selection (Figures~\ref{fig:app11_robustness} and
\ref{fig:app12_selective}), seeded input-order permutations
(Table~\ref{tab:appendix_order_intervention}), budget and profile-aggregation
traces (Table~\ref{tab:appendix_mechanism_trace}), and comparison-set-size
effects (Figure~\ref{fig:app_context_size} and Tables~\ref{tab:context_size_counts}
and~\ref{tab:context_size_focal}).

\section{Conclusion}
Transaction prices offer observable behavioral signals for visual preference,
but non-visual market factors and evaluation protocols complicate their
interpretation. MarketImage tests their suitability through MarketImageArt, a
benchmark pairing prices with human rankings; MarketImageLens, a tie-preserving
diagnostic; and MarketValuator, a comparison of direct, budgeted, and multi-agent
interfaces. Only 24.2\% of pairs have a non-tied price direction, and 34.7\%
of these agree with the human
majority. On this protocol-supported subset, HPSv2 leads at 66.4\% Recoverable
Accuracy; GPT-5.5 Pair and GPT-5.6 Sol Pair reach 60.3\% and 62.1\%. Budgeted
and aggregated interfaces lower Qwen recovery, while GPT gains remain
statistically unresolved. Thus, price directions are sparse,
weakly corroborated, and order-confounded; they are not validated ground truth
for visual preference. Future work should test broader markets with randomized
presentation.

\clearpage
\bibliographystyle{ACM-Reference-Format}
\bibliography{sample-base}

\appendix
\renewcommand{\arraystretch}{0.6}
\section{Reference-Signal and Evaluator Robustness Audit}
\label{sec:appendix}
\label{app:reference_order}
\label{app:evaluation_robustness}
\label{app:order_intervention}

We compare the released three-of-four consensus with all four complete
rankings on the same 266 sets, retaining 2:2 votes as ties. We keep the
released consensus as primary because its exclusion rationale was not recorded;
selected-three agreement statistics appear in the main text.

\begin{figure*}[t]
\centering
\scalebox{1}[0.6]{%
\includegraphics[width=0.99\textwidth]{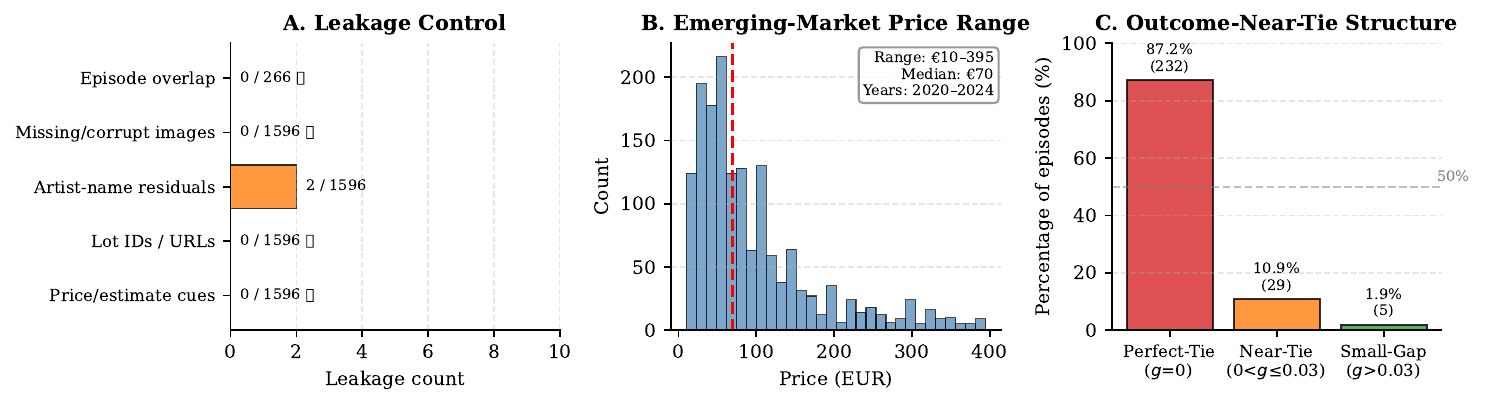}
}
\caption{
Benchmark validation for MarketImageArt.
Left: leakage control confirms that evaluator inputs are free of explicit price-related cues.
Middle: the retained emerging-art records span EUR 10--395 (median EUR 70).
Right: the canonical comparison sets are dominated by zero top-two price gaps.
The implicit price information in source order is audited separately.
}
\Description{Three-panel benchmark audit showing input leakage controls, transaction-price distribution, and the frequency of tied top-two prices.}
\label{fig:emergingart_validation}
\end{figure*}

\begin{table}[t]
\centering
\small
\renewcommand{\arraystretch}{0.85}
\setlength{\tabcolsep}{3pt}
\caption{Post-hoc annotator-count sensitivity, comparing the released three-file consensus with all four complete files. The all-four analysis retains 2:2 votes as ties. Price--human agreement conditions on price-ordered pairs with decisive human votes (966 for three annotators; 753 for four); recovery uses each condition's supported subset and the same saved model outputs.}
\label{tab:appendix_annotator_sensitivity}
\begin{tabular}{p{0.48\columnwidth}rr}
\toprule
\textbf{Measure} & \textbf{Selected 3} & \textbf{All 4} \\
\midrule
Price-ordered pairs & 966 & 966 \\
Human ties & 0 & 213 \\
Decisive human pairs & 966 & 753 \\
Price--human supported pairs & 335 & 257 \\
Price--human conflicts & 631 & 496 \\
Price--human agreement (\%) & 34.7 & 34.1 \\
\midrule
\multicolumn{3}{l}{\textit{Recoverable Accuracy (\%)}} \\
HPSv2 & 66.4 & 66.5 \\
Qwen Pair & 8.4 & 8.2 \\
GPT-5.5 Pair & 60.3 & 61.5 \\
GPT-5.6 Pair & 62.1 & 62.3 \\
\bottomrule
\end{tabular}
\end{table}

The all-four analysis preserves the qualitative recovery ordering despite a
smaller supported subset. As reported in the main text, 154 sets have all
prices tied; among the remaining 112, 71 contain two distinct prices, 25
three, 14 four, and two five, with none containing six.

To test whether larger price differences are more human-corroborated, we
stratify the 966 price-ordered pairs by absolute EUR gap and by relative gap
$|p_i-p_j|/\max(p_i,p_j)$. Absolute-gap bins show little change in hard
agreement (34.2--37.2\%); relative-gap results are shown below:

\begin{table}[t]
\centering
\small
\setlength{\tabcolsep}{3pt}
\caption{Reference reliability by relative price gap $|p_i-p_j|/\max(p_i,p_j)$ among price-ordered pairs. Relative normalization reveals a condition hidden by absolute EUR bins, while the largest-gap bin remains small.}
\label{tab:appendix_relative_gap}
\resizebox{\columnwidth}{!}{%
\begin{tabular}{lrrrr}
\toprule
\textbf{Relative gap} & \textbf{Pairs} & \textbf{Supported} & \textbf{Hard agree [95\% CI]} & \textbf{Soft agree} \\
\midrule
$\leq$1\% & 225 & 63 & 28.0 [19.6, 36.3] & 31.1 \\
1--2\% & 364 & 122 & 33.5 [27.5, 39.8] & 35.7 \\
2--5\% & 323 & 125 & 38.7 [30.4, 46.9] & 41.1 \\
$>$5\% & 54 & 25 & 46.3 [30.6, 63.2] & 47.5 \\
\bottomrule
\end{tabular}%
}
\end{table}

Agreement rises descriptively with relative gap, but the largest-gap bin has
only 54 pairs; this trend does not establish price as a reliable visual
reference. The fixed-order evidence is reported in Benchmark Validation and
RQ1/RQ2. The following checks test whether evaluator conclusions depend on
scoring choices, output margins, or presentation order.

\begin{figure*}[t]
\centering
\includegraphics[width=0.99\textwidth]{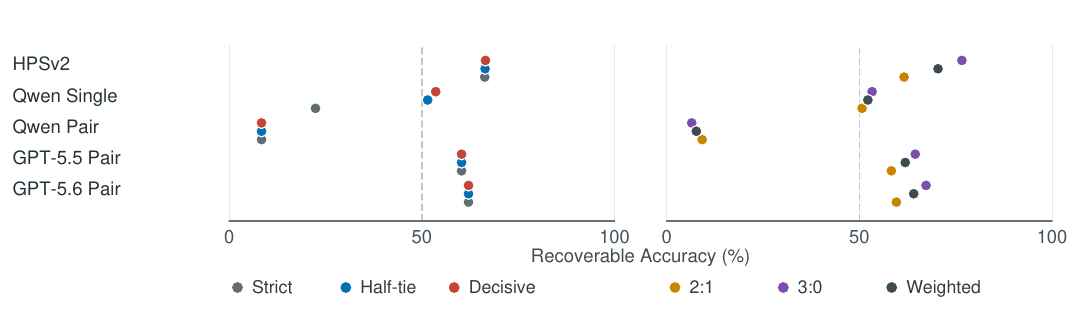}
\caption{Recoverable Accuracy under strict, half-tie, and decisive scoring,
alongside accuracy stratified by 2:1, 3:0, and consensus-weighted human
judgments. The five methods are defined in the plot; values are percentages.}
\Description{Dot plot compares evaluator recovery under alternate prediction-tie scoring and human-consensus definitions.}
\label{fig:app11_robustness}
\end{figure*}

\begin{figure*}[t]
\centering
\scalebox{1}[0.7]{\includegraphics[width=0.86\textwidth]{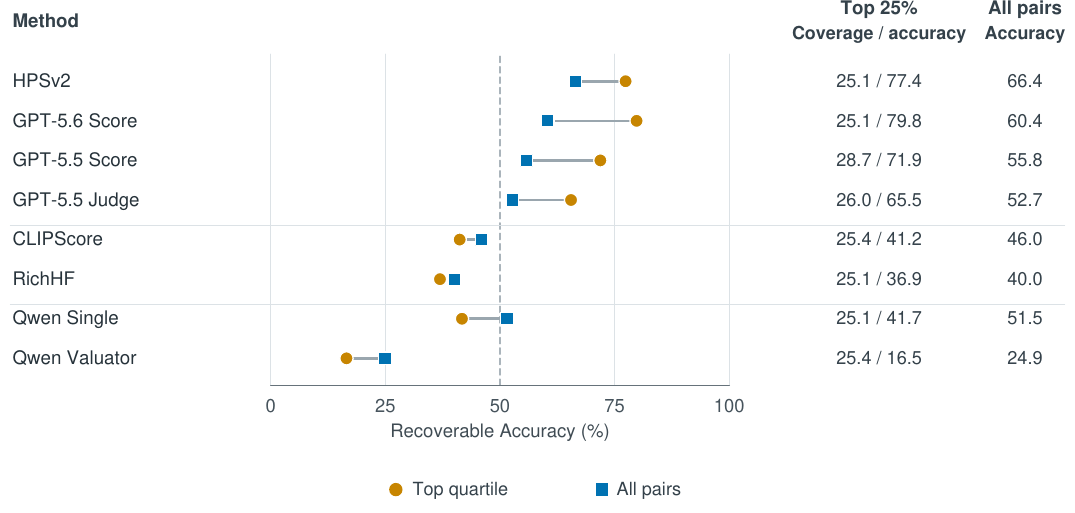}}
\caption{Within-method recovery for the highest-margin nominal quartile and
all 335 supported pairs. Circles denote selected pairs and squares all pairs;
boundary ties are retained, so realized coverage may exceed 25\%. Short method
names are defined in the plot; values are percentages.}
\Description{Selective recovery plot compares high-margin subsets with all supported pairs for visual preference evaluators.}
\label{fig:app12_selective}
\end{figure*}

On the 335 supported pairs, we vary prediction-tie scoring and human-consensus
definitions (Figure~\ref{fig:app11_robustness}). Most static and API results
are stable; Qwen Single-agent is sensitive because it predicts ties on many pairs
(22.4\% strict versus 51.5\% half-credit and 53.6\% decisive-only). HPSv2
remains near 66\% overall and reaches 76.6\% on unanimous pairs, above the
GPT Pair interfaces (64.5\% and 67.3\%). We also rank pairs within each method
by $|\Delta z|$ and compare the highest-margin nominal quartile with all
supported pairs; boundary ties are retained and realized coverage is reported
(Figure~\ref{fig:app12_selective}). HPSv2 rises from 66.4\% to 77.4\% at 25.1\%
coverage, but margins are not calibrated confidence and gains vary by method.

To estimate model-side order sensitivity, we rerun GPT-5.5 and GPT-5.6 Sol
Pair on a fixed subset of 50 canonical sets under three seeded permutations
(22, 42, 168). Outputs are mapped back to item IDs before scoring, yielding
300 complete model--set runs.

\begin{table}[t]
\centering
\small
\renewcommand{\arraystretch}{0.4}
\setlength{\tabcolsep}{3pt}
\caption{Seeded input-order intervention on 50 sets. Recovery ranges span the three permutations; pair agreement and unanimity use the 49 sets with distinct orders. Earlier is the percentage of decisive predictions favoring the earlier attachment.}
\label{tab:appendix_order_intervention}
\resizebox{\columnwidth}{!}{%
\begin{tabular}{lrrrr}
\toprule
\textbf{Model} & \textbf{Rec. Acc. range} & \textbf{Pair agreement} & \textbf{3-order unanimity} & \textbf{Earlier range} \\
\midrule
GPT-5.5 & 53.6--55.4 & 83.3 & 75.0 & 42.7--50.5 \\
GPT-5.6 Sol & 57.1--60.7 & 84.2 & 76.3 & 49.1--51.3 \\
\bottomrule
\end{tabular}%
}
\end{table}

Recovery varies little across permutations, and neither model shows a
consistent attachment-position preference. Because only the two GPT Pair
systems are rerun, this experiment does not resolve the human-order confound or
test Qwen Pair.

\section{Market Elicitation and Mechanism Diagnostics}
\label{app:mechanism_diagnostics}
To diagnose the budget and aggregation results, we inspect budget compliance,
agreement among individual profiles, and sensitivity to the aggregation
reducer.

\subsection{Budget Use, Agent Disagreement, and Aggregation}
\label{app:mechanism_trace}
We analyze API-native allocations and three individual judge or bidder vectors
for GPT-5.5 and GPT-5.6 Sol; legacy Qwen files lack aligned per-profile
vectors. Table~\ref{tab:appendix_mechanism_trace} summarizes budget use,
allocation concentration, and profile agreement.

\begin{table}[t]
\centering
\scriptsize
\renewcommand{\arraystretch}{0.6}
\setlength{\tabcolsep}{2.5pt}
\caption{Audit of API-native budget and multi-agent traces. Budget is mean budget utilization, Full is the fraction of vectors exactly exhausting the cap, and Top is the largest item's mean allocation share. Agent agr. is exact pair-direction agreement across profile pairs. Unanim. is the share of recoverable pairs on which all three profiles agree; Acc. U/N gives aggregate accuracy on unanimous/non-unanimous pairs. Dashes denote inapplicable quantities.}
\label{tab:appendix_mechanism_trace}
\resizebox{\columnwidth}{!}{%
\begin{tabular}{llrrrrrr}
\toprule
\textbf{Model} & \textbf{Interface} & \textbf{Budget} & \textbf{Full} & \textbf{Top} & \textbf{Agent agr.} & \textbf{Unanim.} & \textbf{Acc. U/N} \\
\midrule
GPT-5.5 & Budgeted Single-agent & 99.9 & 99.2 & 27.3 & -- & -- & -- \\
 & Valuator Judge & -- & -- & -- & 73.8 & 58.5 & 56.1/47.8 \\
 & Valuator & 99.9 & 98.5 & 27.2 & 72.3 & 57.3 & 53.1/50.3 \\
\midrule
GPT-5.6 Sol & Budgeted Single-agent & 100.0 & 100.0 & 26.7 & -- & -- & -- \\
 & Valuator Judge & -- & -- & -- & 78.8 & 66.0 & 57.9/50.0 \\
 & Valuator & 100.0 & 98.4 & 26.3 & 74.8 & 57.0 & 59.2/55.6 \\
\bottomrule
\end{tabular}%
}
\end{table}

Budget use is near-complete (99.9--100.0\% mean utilization), so cap
under-spending does not explain the lack of a consistent mechanism gain.
Profiles agree on 72.3--78.8\% of pair directions, while unanimity covers
57.0--66.0\% of recoverable pairs. Unanimous-profile accuracy is higher in
all four model/interface rows, most clearly for GPT-5.5 Valuator Judge
(56.1\% versus 47.8\%); this descriptive split is not an independent
significance test. Alternative reducers yield 51.3--52.5\% recovery for
GPT-5.5 and 54.6--57.6\% for GPT-5.6 Sol, with no consistent post-hoc gain.

\section{Supplementary Analysis Notes}
The reference decomposition and three diagnostic questions were fixed before
the full API suites. Gap bins, alternative tie scoring, margin selection,
aggregation variants, and the order intervention are descriptive robustness
analyses, not multiple-comparison-corrected findings. Bootstrap intervals use
seed 2027 with deterministic model--method offsets; the order intervention
uses seeds 22, 42, and 168. Metric definitions, denominators, and run-completion
checks are given in the main Experimental Setup.

\section{Matched Comparison-Set Size Ablation}
\label{app:context_size}
\label{app:reproducibility_scope}

\begin{figure*}[t]
\centering
\begin{minipage}[t]{0.49\textwidth}
\centering
\includegraphics[width=\linewidth]{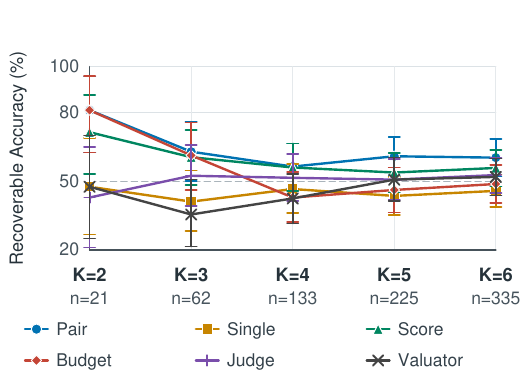}
\end{minipage}\hfill
\begin{minipage}[t]{0.49\textwidth}
\centering
\includegraphics[width=\linewidth]{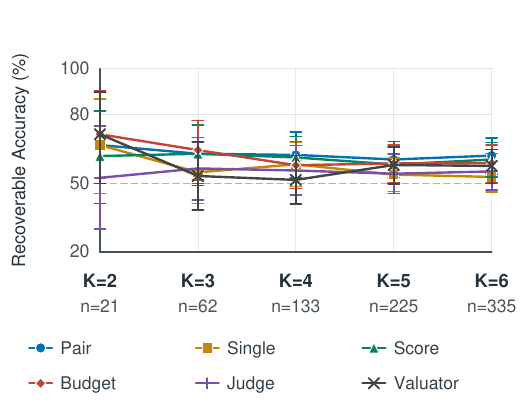}
\end{minipage}
\caption{Recoverable Accuracy in the matched comparison-set-size ablation.
Left: GPT-5.5; right: GPT-5.6 Sol. Every condition contains 266 sets, and
$n$ denotes recoverable pairs. Whiskers are 95\% set-cluster bootstrap
intervals; the dashed line marks 50\%. The vertical axis spans 20--100\%.
Legend labels Pair, Single, Score, Budget, and Judge denote Single-agent
(Pair), Single-agent, Single-agent (LLM Score), Budgeted Single-agent, and
Valuator Judge, respectively; Valuator keeps its full name.
The trajectories are descriptive and do not test differences between set
sizes.}
\Description{Side-by-side line plots show Recoverable Accuracy and
95-percent set-cluster intervals across comparison-set sizes for GPT-5.5 and
GPT-5.6 Sol.}
\label{fig:app_context_size}
\end{figure*}

\subsection{Question and Matched Data Construction}
We test whether joint-prompt context size changes the six GPT elicitation
interfaces. From each canonical six-image set, deterministic SHA-256
priorities (seed 2027, set ID, item ID) define nested subsets for
$K\in\{2,3,4,5,6\}$. Selection uses no prices, human votes, or model outputs;
selected items retain their canonical relative order. Thus, all conditions
contain 266 sets but have different image and pair counts. For set $s$:
\begin{equation}
S_s^{(2)}\subset S_s^{(3)}\subset S_s^{(4)}
\subset S_s^{(5)}\subset S_s^{(6)}.
\end{equation}
The same subsets are used for both models and all six methods.
Table~\ref{tab:context_size_counts} reports the resulting reference counts.

\begin{table}[t]
\centering
\scriptsize
\renewcommand{\arraystretch}{0.7}
\setlength{\tabcolsep}{2pt}
\caption{Matched set-size conditions (266 sets each). Pairs counts all within-set comparisons; Price-ord. counts non-tied pairs, split into Supported and Conflicts; Ties are price-tied pairs.}
\label{tab:context_size_counts}
\begin{tabular}{rrrrrrr}
\toprule
$K$ & Images & Pairs & Supported & Price-ord. & Conflict & Ties \\
\midrule
2 & 532 & 266 & 21 & 65 & 44 & 201 \\
3 & 798 & 798 & 62 & 198 & 136 & 600 \\
4 & 1,064 & 1,596 & 133 & 410 & 277 & 1,186 \\
5 & 1,330 & 2,660 & 225 & 660 & 435 & 2,000 \\
6 & 1,596 & 3,990 & 335 & 966 & 631 & 3,024 \\
\bottomrule
\end{tabular}
\end{table}
\begin{table}[H]
\centering
\small
\renewcommand{\arraystretch}{0.4}
\setlength{\tabcolsep}{3pt}
\caption{Change in Human Acc.\ ($K=6$ minus $K=2$, percentage points) on the same 266 focal pairs; ties receive half credit. Method names are abbreviated as in Table~\ref{tab:pairwise_diagnosis}. Per-image scores are reused; joint interface changes are descriptive, not significance tests.}
\label{tab:context_size_focal}
\begin{tabular}{lrr}
\toprule
\textbf{Method} & \textbf{GPT-5.5} & \textbf{GPT-5.6 Sol} \\
\midrule
Pair & -5.6 & -1.9 \\
Single & 0.0 & 0.0 \\
Score & 0.0 & 0.0 \\
Budget & -7.1 & -4.1 \\
Judge & -13.0 & -4.7 \\
Valuator & -16.5 & -6.6 \\
\bottomrule
\end{tabular}
\end{table}
Human rankings are restricted from the original six-image annotations, not
recollected at smaller $K$; they therefore represent six-image preferences.
Retaining canonical order also preserves the price-order exposure audited in
Appendix~\ref{app:reference_order}.

\subsection{Interfaces and Execution Accounting}
GPT-5.5 and GPT-5.6 Sol use medium reasoning effort and the six interfaces in
Table~\ref{tab:pairwise_diagnosis}. Joint-interface prompts and output lengths
are resized to $K$, with personas and aggregation unchanged. Budgeted methods
receive EUR~$100K$ per image, a constant-per-image rather than fixed-total
budget.

The four joint interfaces are rerun for each smaller $K$. The two image-wise
methods reuse saved $K=6$ scores projected onto selected items; their pair
metrics can change with pair composition, but not with comparison context.

The matrix contains 48 files and 12,768 method--set outputs. Its 32 joint
interface files add 8,512 set outputs and require
\begin{equation}
2\times4\times266\times(1+1+3+3)=17{,}024
\end{equation}
logical calls. Traces record 17,025 attempts, including one successful retry,
with no final failure or parser fallback. The other 16 files contain 4,256
projected outputs and require no calls. All outputs pass coverage, identity,
order, cardinality, and finite-value checks.

\subsection{All Within-Set Pair Results}
Across all within-set pairs, Figure~\ref{fig:app_context_size} shows no
uniform monotonic trend; changing pair identities and supported-pair counts
prevent attributing differences to context size alone. On the same 266 focal
pairs, joint-interface Human Accuracy falls by 1.9--16.5 points from $K=2$ to
$K=6$, while per-image methods are unchanged because their scores are reused
(Table~\ref{tab:context_size_focal}). This descriptive probe is not a
significance test.

\subsection{Fixed-Pair Comparison Across Context Sizes}
To isolate context from pair composition, we track the same $K=2$ image pair
within each of 266 sets (21 recoverable, 44 conflicting, 201 price-tied).
Table~\ref{tab:context_size_focal} reports the change in Human Accuracy at
$K=6$ relative to $K=2$.

\section{Limitations and Ethics.} 
(1) MarketImageArt covers one
low-price emerging-art source and constructed comparison sets, not simultaneous
auctions. (2) Human judgments used a fixed price-sorted order, and the released
consensus selected three of four complete annotation files without a documented
rationale; causal claims about the visual contribution to price are therefore
unsupported. (3) GPT and Qwen interfaces differ in prompts, call graphs, and
compute, so cross-family comparisons are system-level. (4) Human rankings are
used only in aggregate; annotator demographics, recruitment, and compensation
are undocumented. Generated prices and bids are not appraisals and should not
guide purchases or claims about artists; any data release must respect image
licenses and privacy and distinguish market outcomes from human judgments and
model predictions.
\end{document}